\documentclass[sigconf]{acmart}
\AtBeginDocument{%
  }

\usepackage{multirow}
\usepackage{enumitem}
\usepackage{colortbl}
\usepackage{tablefootnote}
\usepackage{makecell}
\usepackage{balance}
\usepackage{fontawesome}
\usepackage{algorithm}
\usepackage{algorithmic}

\usepackage{fancyvrb}
\usepackage{xcolor}

\newcommand{\ie}{\emph{i.e., }}

\copyrightyear{2026}
\acmYear{2026}
\setcopyright{cc}
\setcctype{by}
\acmConference[WWW '26] {Proceedings of the ACM Web Conference 2026}{April 13--17, 2026}{Dubai, United Arab Emirates.}
\acmBooktitle{Proceedings of the ACM Web Conference 2026 (WWW '26), April 13--17, 2026, Dubai, United Arab Emirates}
\acmISBN{979-8-4007-2307-0/2026/04}
\acmDOI{10.1145/10.1145/3774904.3792905}

\begin{document}

\title{ThinkTank-ME: A Multi-Expert Framework for \\Middle East Event Forecasting}

\author{Haoxuan Li}
\affiliation{%
  \institution{University of Electronic Science and Technology of China}
  \city{Chengdu}
  \country{China}
}
\email{lhx980610@gmail.com}

\author{He Chang}
\affiliation{%
  \institution{Communication University of China}
  \city{Beijing}
  \country{China}
}
\email{hechangcuc@cuc.edu.cn}

\author{Yunshan Ma}
\authornote{Corresponding author}
\affiliation{%
  \institution{Singapore Management University}
  \city{Singapore}
  \country{Singapore}
}
\email{ysma@smu.edu.sg}

\author{Yi Bin}
\affiliation{%
  \institution{Tongji University}
  \city{Shanghai}
  \country{China}
}
\email{yi.bin@hotmail.com}

\author{Yang Yang}
\affiliation{%
  \institution{University of Electronic Science and Technology of China}
  \city{Chengdu}
  \country{China}
}
\email{yang.yang@uestc.edu.cn}

\author{See-Kiong Ng}
\author{Tat-Seng Chua}
\affiliation{%
  \institution{National University of Singapore}
  \city{Singapore}
  \country{Singapore}
}
\email{seekiong@nus.edu.sg}
\email{dcscts@nus.edu.sg}

\renewcommand{\shortauthors}{Haoxuan Li et al.}

\begin{abstract}
  Event forecasting is inherently influenced by multifaceted considerations, including international relations, regional historical dynamics, and cultural contexts. 
  However, existing LLM-based approaches employ single-model architectures that generate predictions along a singular explicit trajectory, constraining their ability to capture diverse geopolitical nuances across complex regional contexts. 
  To address this limitation, we introduce \textbf{ThinkTank-ME}, a novel \textbf{Think} \textbf{Tank} framework for \textbf{M}iddle \textbf{E}ast event forecasting that emulates collaborative expert analysis in real-world strategic decision-making. 
  To facilitate expert specialization and rigorous evaluation, we construct \textbf{POLECAT-FOR-ME}, a Middle East–focused event forecasting benchmark.
  Experimental results demonstrate the superiority of multi-expert collaboration in handling complex temporal geopolitical forecasting tasks. The code is available at \textcolor{magenta}{\url{https://github.com/LuminosityX/ThinkTank-ME}}.
\end{abstract}




\maketitle

\section{Introduction}








Event forecasting aims to predict future events from historical observations. 
Accurate and reliable event forecasting is critical for proactive decision-making, which enables optimized resource allocation, early risk mitigation, and reduced societal disruptions~\cite{zhao2021event}. Given its profound implications, event forecasting~\cite{Foundation_TS, SeCoGD} has garnered increasing attention across both academia and industry.

\begin{figure}[t]
\centering
\includegraphics[width=0.9\linewidth]{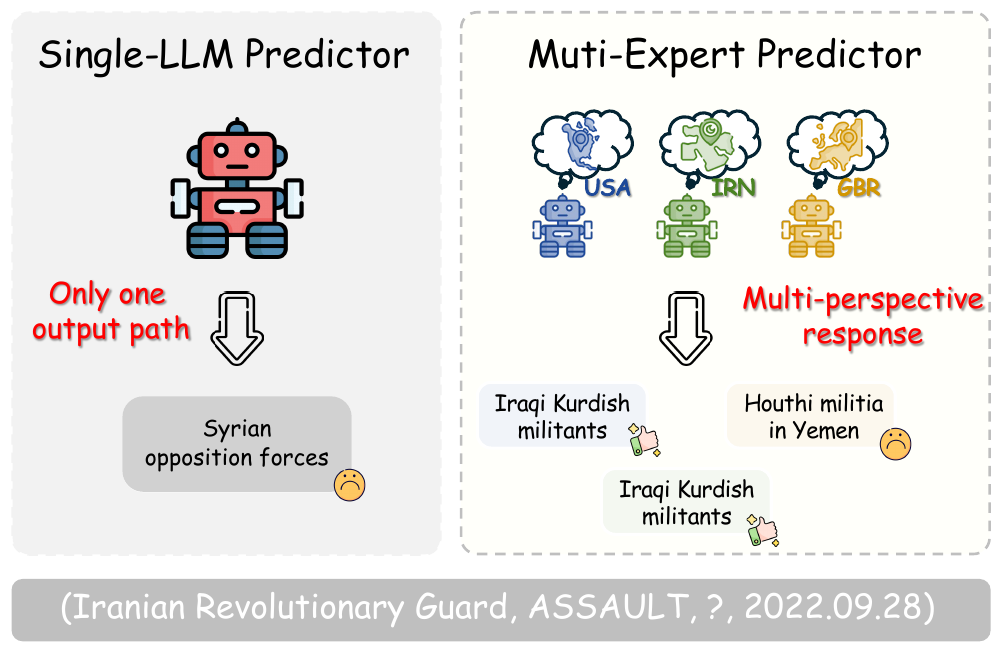}
\caption{Illustration of the insufficiency of relying solely on a single LLM to analyze complex real-world events. 
}
\label{fig:fig_1}
\end{figure}

With the AGI wave led by Large Language Models (LLMs), leveraging LLMs for event forecasting~\cite{GENTKG, Chain-of-History, halawi2024approaching, deng2024advances} has emerged as a promising direction. The inherent capability of LLMs~\cite{hurst2024gpt, grattafiori2024llama, shi2024math} to process vast amounts of text makes them particularly well-suited for event forecasting tasks, as event sequences are typically accompanied by rich textual information. 
Recent pioneering studies~\cite{GENTKG, li2024mm} have explored the application of LLMs in event forecasting by employing prompting strategies such as in-context learning~\cite{zhang2024analyzing} and Chain-of-Thought~\cite{Chain-of-History}.
These advances underscore the potential of LLMs in redefining event forecasting methodologies, paving the way for more robust and superior predictive models. 

\begin{figure*}[!ht]
\centering
\includegraphics[width=\textwidth]{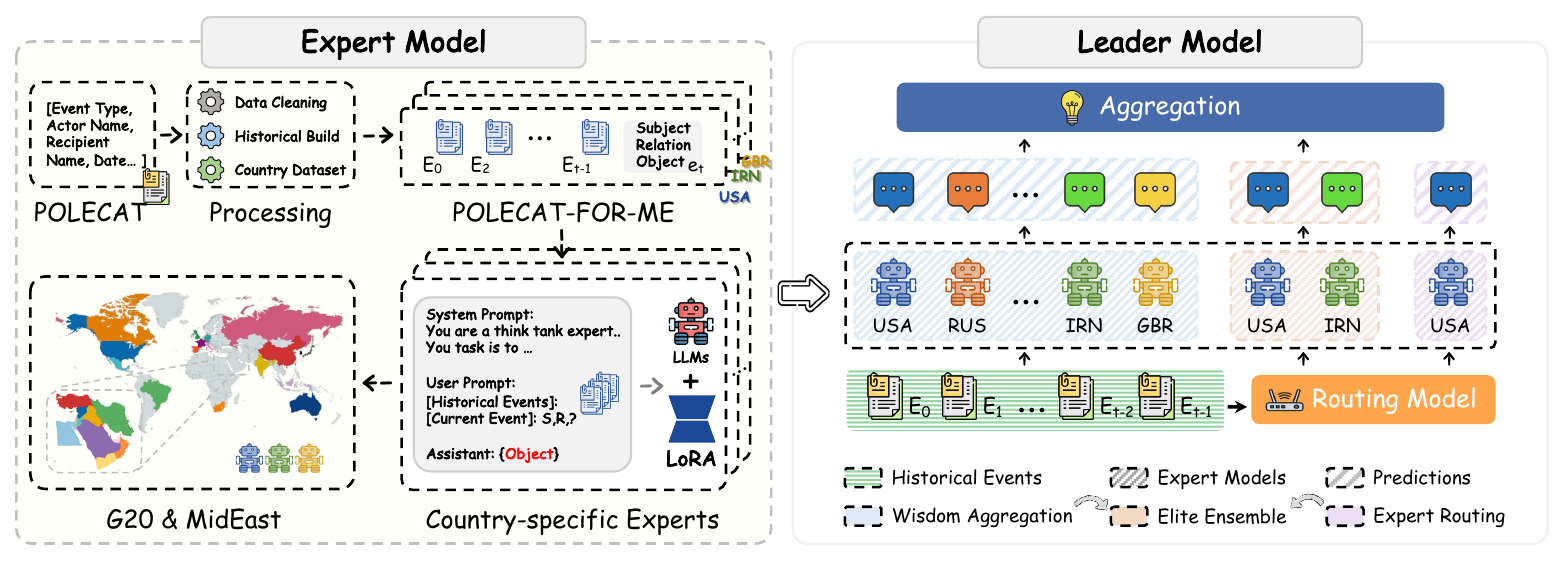}
\caption{Schematic overview of the proposed ThinkTank-ME framework. This framework comprises multiple expert models, each specializing in a specific domain, coordinated by a leader model that synthesizes their diverse predictions. For the leader model, we investigate three distinct aggregation strategies: Expert Routing, Wisdom Aggregation, and Elite Ensemble.}
\label{fig:fig_2}
\vspace{+0.1in}
\end{figure*}

Despite these advances, existing LLM-based approaches encounter notable limitations when applied to complex real-world scenarios.
Real-world events are influenced by multifaceted factors, such as geopolitical interests, cultural contexts, and economic conditions. 
Considering these factors from different perspectives or stakeholder positions yields divergent predictions.
Recent studies~\cite{chen2025reasoning, hao2024training} have demonstrated that a single LLM often implicitly adopts a dominant perspective or reasoning pattern, with its outputs typically following one explicit trajectory, as illustrated in Figure~\ref{fig:fig_1}.
Consequently, single-LLM predictors struggle to capture nuanced regional characteristics and intricate dynamic relationships, ultimately limiting their potential to achieve high-precision forecasting in complex real-world scenarios.


To address this challenge, inspired by the operational mechanisms of real-world think tanks~\cite{pautz2011revisiting}, we propose \textbf{ThinkTank-ME}, a novel \textbf{Think} \textbf{Tank} framework for \textbf{M}iddle \textbf{E}ast event forecasting. 
Our framework integrates multiple specialized expert models, each targeting specific regional domains, coordinated by a leader model that synthesizes their collective predictions.
Specifically, we construct expert models with targeted expertise through parameter-efficient fine-tuning on country-level event datasets. 
For the leader model, we explore various information aggregation strategies~\cite{schoenegger2024wisdom, saha2025wisdom, mannes2014wisdom} to synthesize predictions from individual experts. 
Equally importantly, recognizing the insufficient granularity of expert specialization in existing datasets~\cite{ICEWS, GDELT, halterman2023plover} and the unique international influence and high complexity of events in the Middle East, we construct the POLECAT-FOR-ME dataset.
Experimental results demonstrate the superiority of multi-expert collaboration in handling complex temporal geopolitical event forecasting tasks.
The main contributions are as follows:
\begin{itemize}[leftmargin=*]
    \item To the best of our knowledge, we are the first to introduce a ``think tank'' simulation approach for event forecasting, establishing a novel paradigm for multi-expert collaborative prediction. 
    \item We propose ThinkTank-ME, an innovative framework that integrates multiple specialized expert models coordinated by a leader model to enhance forecasting accuracy.
    \item To address the limitations of existing datasets and to facilitate research on complex geopolitical events in the Middle East, we construct a new benchmark dataset, POLECAT-FOR-ME.
    \item Experimental results on POLECAT-FOR-ME validate the effectiveness of the proposed ThinkTank-ME framework.
\end{itemize}

\section{Problem Formulation}

Following prior works, we represent each temporal event as a structured quadruple $(s, r, o, t)$, where $s \in \mathcal{E}$ denotes the subject entity, $r \in \mathcal{R}$ denotes the relation, $o \in \mathcal{E}$ denotes the object entity, and $t$ denotes the timestamp. Here, $\mathcal{E}$ and $\mathcal{R}$ represent the sets of all entities and relations, respectively. At a given timestamp $t$, the set of observed events is denoted as $E_t = \{(s_i, r_i, o_i, t_i)\}_{i=1}^{N}$, where $N$ is the number of events observed at time $t$. Given the historical event sequences up to time $t$, \ie $E_{<t} = \{E_1, E_2, \dots, E_{t-1}\}$, and a forecasting query at time $t$, typically in the form of a partially specified triple $(s, r, ?)$, the goal is to predict the missing object entity $o$.

\section{ThinkTank-ME}


\begin{table*}[!ht]
\centering
\caption{Performance (accuracy) comparison across baseline methods, proprietary LLMs, and ThinkTank-ME.
For country-specific accuracy, we report results on representative countries from the Middle East. The best results among non-proprietary models are marked in bold, while the overall best results are underlined. Mi and Ma denote micro- and macro-average accuracy.
}
\label{tab:tab1}
\begin{tabular}{ll||cccccccc||cc}
\specialrule{1.0pt}{0pt}{0pt}
\multirow{2}{*}{\textbf{Methods}} & \multirow{2}{*}{\textbf{Variants}} & \multicolumn{8}{c||}{Country-specific Accuracy} & \multicolumn{2}{c}{Total Accuracy} \\
\cmidrule{3-10}
\cmidrule{11-12}
& & {\textbf{ISR}} & {\textbf{EGY}} & {\textbf{IRQ}} & {\textbf{YEM}} & {\textbf{SYR}} & {\textbf{LBN}} & {\textbf{QAT}} & {\textbf{PSE}} & {\textbf{Mi}} & {\textbf{Ma}} \\
\midrule                     
\multirow{2}{*}{Baselines} & {No Training} & {11.1} & {7.8} & {9.6} & {34.0} & {17.8} & {12.4} & {10.1} & {8.3} & {15.5} & {15.7} \\
& {All-data Training} & {17.4} & {10.7} & {16.2} & {49.9} & {19.3} & {25.3} & {8.5} & {15.8} & {22.7} & {22.6}\\
\midrule
\multirow{2}{*}{Proprietary LLMs} & {GPT-4o~\cite{hurst2024gpt}} & {\underline{22.2}} & {17.3} & {\underline{15.3}} & {37.9} & {18.6} & {15.6} & {\underline{20.8}} & {15.8} & {23.5} & {24.0} \\
& {GPT-4o-mini~\cite{hurst2024gpt}} & {20.3} & {\underline{21.3}} & {15.3} & {37.2} & {17.7} & {16.3} & {19.2} & {14.1} & {22.7} & {22.5}\\
\midrule
{Expert Routing} & - & {13.9} & {8.4} & {11.4} & {39.0} & {15.6} & {23.7} & {10.7} & {13.5} & {18.6} & {19.4}\\
\midrule
\multirow{3}{*}{Wisdom Aggregation} & {Majority Voting} & {17.7} & {11.9} & {13.9} & {50.4} & {21.6} & {26.9} & {11.2} & {18.1} & {23.6} & {23.9}\\
& {Vanilla Best-of-N} & {16.3} & {13.1} & {9.6} & {48.4} & {18.2} & {19.6} & {9.0} & {15.5} & {21.2} & {22.2}\\
& {Weighted Best-of-N} & {18.0} & {13.9} & {14.4} & {50.7} & {22.2} & {26.6} & {10.9} & {17.5} & {23.9} & {24.4}\\
\midrule
\multirow{3}{*}{Elite Ensemble} & {Majority Voting} & {18.0} & {12.8} & {13.5} & {50.7} & {22.6} & {26.9} & {11.2} & {17.0} & {23.9} & {24.2}\\
& {Vanilla Best-of-N} & {17.7} & {14.1} & {11.8} & {50.9} & {19.2} & {24.8} & {8.0} & {16.1} & {22.8} & {23.9}\\
& {Weighted Best-of-N} & {\textbf{18.3}} & {\textbf{15.7}} & {\textbf{14.4}} & {\textbf{\underline{51.6}}} & {\textbf{\underline{23.4}}} & {\textbf{\underline{26.9}}} & {\textbf{11.5}} & {\textbf{\underline{17.5}}} & {\textbf{\underline{24.6}}} & {\textbf{\underline{25.0}}}\\
\specialrule{1.0pt}{0pt}{0pt}
\end{tabular}
\vspace{+0.1in}
\end{table*}

The overall architecture of ThinkTank-ME is illustrated in Figure~\ref{fig:fig_2}, comprising expert models and a leader model.

\footnotetext[1]{Middle East: IRN, ISR, EGY, SAU, TUR, IRQ, YEM, SYR, JOR, ARE, LBN, OMN, KWT, QAT, BHR, CYP, PSE. Global (G20): CHN, USA, RUS, GBR, FRA, DEU, KOR, JPN, IND, CAN, ITA, AUS, ESP, ARG, BRA, IDN, MEX, ZAF. Country codes based on ISO 3166.}

\subsection{Dataset Construction}
\label{sec:3_1}
Standard event forecasting datasets such as ICEWS~\cite{ICEWS}, GDELT~\cite{GDELT} rely on the CAMEO ontology~\cite{CAMEO}, whose 2009 schema struggles to reflect modern geopolitical complexity. To overcome these limitations, we first adopt the POLECAT dataset~\cite{halterman2023plover}, which provides finer temporal–spatial granularity and broader event coverage enabled by modern AI-driven event extraction techniques. Subsequently, as illustrated in Figure~\ref{fig:fig_2}, we construct POLECAT-FOR-ME, a specialized dataset curated for Middle East event forecasting through three stages: (i) \textbf{Data cleaning}, including event de-duplication and entity validation to ensure consistency and reliability; (ii) \textbf{Historical sequence formation}, where semantically related events are grouped into historical event sequences following~\cite{SeCoGD}; (iii) \textbf{Country dataset construction}, where events are partitioned by their ``Country'' attribute to create country-specific dataset.
Considering geopolitical relevance and data sufficiency, we select 35 countries and regions\footnotemark[1].
Test set partitioning accounts for the knowledge cutoff of Llama 3.1 (December 2023) to prevent leakage. Additional details are available in the code repository.

\subsection{Think Tank Forecasting: Expert Model}
\label{sec:3_2}
Inspired by the collaborative nature of real-world think tanks, we propose a novel forecasting framework, underpinned by LLMs specialized as expert models. Expert model construction involves two primary components: \textbf{Knowledge Acquisition}, wherein LLMs are fine-tuned on country-specific subsets of the POLECAT-FOR-ME to capture unique regional event patterns and dynamics; and \textbf{Prompt Engineering}, where we design a structured prompt to instruct the LLMs to assume the role of a geopolitical expert and predict the missing object based on historical context. By integrating fine-tuned LLMs with tailored prompts, our framework simulates multidisciplinary think-tank expertise for context-aware geopolitical forecasting.
All expert models are built upon Llama-3.1-8B~\cite{grattafiori2024llama}.

\subsection{Think Tank Forecasting: Leader Model}
\label{sec:3_3}

To aggregate forecasts from multiple experts, a leader model integrates their outputs into a final prediction, mirroring the synthesis process of real-world think tanks. We evaluate three leader strategies: Expert Routing, Wisdom Aggregation, and Elite Ensemble.

\subsubsection{Expert Routing.} 
A straightforward aggregation strategy is to directly assign each query to a single expert, analogous to the MoE paradigm~\cite{cai2024survey}, where a gating mechanism routes inputs to specialized sub-models.
To implement this, we train a routing model by sampling events from the POLECAT-FOR-ME training set and collecting predictions from all fine-tuned experts.
Experts that correctly predict the missing object are treated as supervision signals, producing training pairs of queries and their suitable experts. 
The routing model \textit{R} selects the expert $x_i$ as: $x_i=\textit{R}\ (E_{<t},\ \{s_i,r_i,t_i\}),$
and the forecast of $x_i$ serves as the final output. More implementation details are available in the code repository.


\subsubsection{Wisdom Aggregation.}
Inspired by the wisdom of crowd~\cite{schoenegger2024wisdom}, we explore two strategies: Majority Voting and Best-of-N.
\noindent \textbf{Majority voting} serves as a baseline aggregation approach, in which each expert model predicts the missing object independently, and the most frequent prediction is chosen as the final output.
\noindent \textbf{Best-of-N} extends majority voting by incorporating confidence scores, typically derived from the generation probabilities. We investigate two variants: (i) \textbf{Vanilla Best-of-N}, which selects the prediction with the highest individual confidence; and (ii) \textbf{Weighted Best-of-N}, which sums confidence for identical predictions and selects the candidate with the highest aggregated score. Formally,
\begin{equation}
    S_{agg}(o) = \sum^{N}_{j=1} c_j \cdot \mathbb{I}(o_j = o),
\end{equation}
where $N$ is the number of experts, $o_j$ and $c_j$ are the prediction and confidence score of expert $j$, and $\mathbb{I}$ is the indicator function.


\subsubsection{Elite Ensemble.}
Expert routing and wisdom aggregation offer complementary trade-offs: the former may overlook collective knowledge, whereas the latter requires querying all experts, incurring a higher computational cost. 
To balance accuracy and efficiency, we introduce the Elite Ensemble strategy, which reuses the routing model to rank experts by their predicted probabilities and selects the top-$k$ experts for selective aggregation. 
By leveraging the ``wisdom of the elite'', this approach retains the diversity and accuracy benefits of full aggregation while reducing computational overhead.



\section{Experimental Results}

We conduct comprehensive experiments to evaluate the effectiveness of our proposed ThinkTank-ME framework. The experiments are designed to answer the following research questions:
\begin{itemize}[leftmargin=*]
    \item \textbf{RQ1:} How effective is our proposed ThinkTank-ME framework?
    \item \textbf{RQ2:} Does the country-specific fine-tuning of expert models provide tangible benefits?
    \item \textbf{RQ3:} How do different leader model aggregation strategies within ThinkTank-ME affect the overall performance?
\end{itemize}

\noindent \textbf{RQ1: Framework Effectiveness.} We compare ThinkTank-ME under various leader model strategies against two baselines: \textit{No Training} (vanilla LLM without fine-tuning) and \textit{All-data Training} (single LLM fine-tuned on the complete training dataset). As presented in Table~\ref{tab:tab1}, our ThinkTank-ME framework demonstrates superior performance. Notably, the Elite Ensemble (Weighted Best-of-N) variant achieves micro-average accuracy of 24.6 and macro-average accuracy of 25.0, substantially outperforming \textit{All-data Training} (22.7 Mi, 22.6 Ma) by 8.4\% and 10.6\%, respectively, while significantly surpassing the \textit{No Training} baseline. 
These findings validate the effectiveness of our think tank forecasting paradigm. 
Beyond open-source models, we further compare ThinkTank-ME with two proprietary LLMs, GPT-4o and GPT-4o-mini. As shown in Table~\ref{tab:tab1}, ThinkTank-ME consistently achieves the highest micro- and macro-average accuracy among all models, demonstrating its superiority. Interestingly, for several high-profile countries such as Israel and Egypt, proprietary LLMs exhibit higher country-specific accuracy. We attribute this to the inherent ``long-tail'' distribution of geopolitical data: countries with abundant news coverage and rich linguistic resources are more easily captured by large proprietary models, whereas low-resource countries remain underrepresented. These observations also validate the necessity of expert modeling.
Among aggregation strategies, the elite ensemble strategy constitutes the most effective leader model strategy, as it strategically combines expert routing with the principle of ``wisdom of the crowd''.

\begin{figure}[t]
\centering
\includegraphics[width=\linewidth]{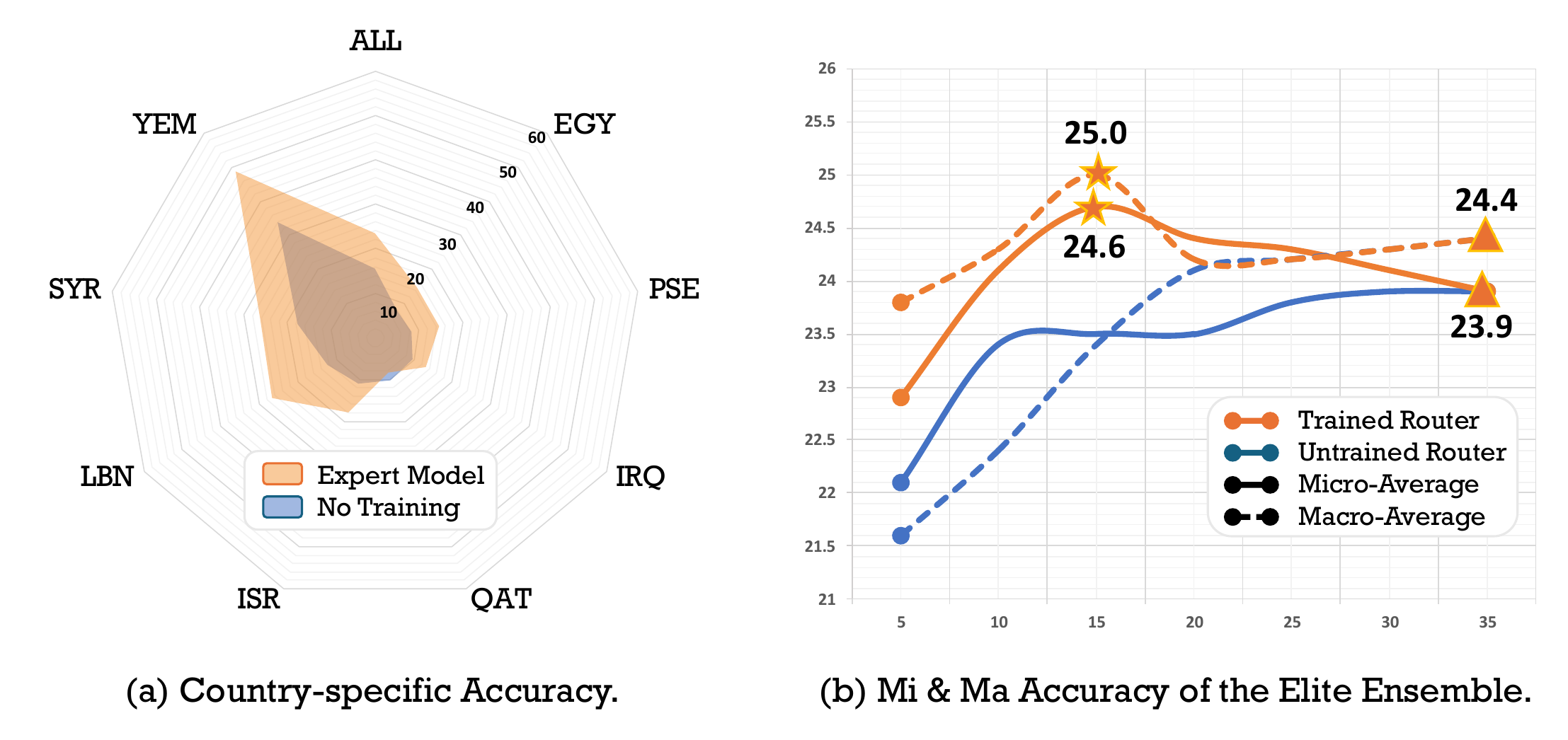}
\caption{Experimental validation of the expert and leader model. 
Subfigure (a) compares the \textit{No Training} baseline with expert models on their country-specific test sets.
Subfigure (b) shows the performance of the Elite Ensemble (Weighted Best-of-N) under varying numbers of selected experts and contrasts it with an untrained router.
The $\star\ \&\ \triangle$ refer to the optimal and the Wisdom Aggregation.
}
\label{fig:fig_3}
\end{figure}

\noindent \textbf{RQ2: Specialized Expert Models.} 
We validate the efficacy of specialized expert models through two complementary analyses. 
Figure~\ref{fig:fig_3}(a) shows that most expert models outperform the \textit{No Training} baseline on their country-specific test sets, indicating that targeted fine-tuning imparts meaningful domain knowledge.
Second, the performance gain of \textit{All-data Training} over \textit{No Training} (Table~\ref{tab:tab1}) further confirms that fine-tuning integrates relevant predictive knowledge into LLMs, even when trained on broader, aggregated dataset. 
These findings collectively support the foundational premise of creating specialized experts through targeted fine-tuning.

\noindent \textbf{RQ3: Leader Model Analysis.} 
Table~\ref{tab:tab1} reveals progressive improvement across aggregation strategies: from Expert Routing (18.6 Mi, 19.4 Ma) to Wisdom Aggregation (23.9 Mi, 24.4 Ma), and to Elite Ensemble (24.6 Mi, 25.0 Ma). This aligns with theoretical expectations: Expert Routing, while computationally efficient, inadequately leverages collective knowledge; Wisdom Aggregation harnesses all experts but can be diluted by irrelevant contributions; Elite Ensemble optimizes this trade-off by selectively combining a filtered group of high-performing experts. To further demonstrate the contribution of the routing mechanism within the elite ensemble strategy, we compare the performance of our trained router against an untrained variant in Figure~\ref{fig:fig_3}(b). The elite ensemble with a trained router exhibits a characteristic performance curve: accuracy generally improves with increasing expert selection until reaching an optimal point, after which incorporating additional irrelevant experts degrades performance. 
This behavior is absent with an untrained router, confirming that the trained router effectively prioritizes proficient experts and optimizes the elite subset.
\section{Conclusion and Future Work}

Inspired by the operational mechanisms of real-world think tanks, we proposed ThinkTank-ME, a novel multi-expert forecasting framework that employs multiple specialized expert models, coordinated by a leader model, to synthesize collective predictions for complex geopolitical event forecasting. By explicitly modeling diverse regional perspectives and expert specializations, the proposed framework mitigates the limitations of single-LLM predictors, which often rely on a single dominant reasoning trajectory and struggle to capture nuanced geopolitical dynamics.
Experimental results on the POLECAT-FOR-ME benchmark demonstrate that ThinkTank-ME consistently outperforms vanilla LLM baselines and single-model fine-tuning strategies under different aggregation schemes, validating the effectiveness of structured multi-expert collaboration for complex temporal forecasting tasks. In particular, the elite ensemble strategy achieves a favorable balance between accuracy and computational efficiency by selectively aggregating high-performing experts.
Future work involves leveraging the reasoning capabilities of LLMs to improve accuracy, and scaling the framework to broader geographical regions to validate its generalizability.
\begin{acks}
This work is partially supported by the National Natural Science Foundation of China under grant 62220106008.
This research was supported by the Singapore Ministry of Education (MOE) Academic Research Fund (AcRF) Tier 1 grant.
\end{acks}


\bibliographystyle{ACM-Reference-Format}
\balance
\bibliography{sample-base}










\end{document}